\documentclass[letterpaper, 10 pt, conference]{ieeeconf}  

\IEEEoverridecommandlockouts                              

\usepackage{amsmath,amssymb,amsfonts}
\usepackage{graphicx}
\usepackage{algorithm}
\usepackage{algpseudocode}
\usepackage{subfig}
\usepackage{overpic}
\usepackage[colorlinks=true, allcolors=blue]{hyperref}
\usepackage{multirow}
\usepackage{threeparttable}
\usepackage{bm}
\newtheorem{assumption}{Assumption}

\newcommand{\mc}[1]{\mathcal{#1}}

\title{\LARGE \bf
Five-Tiered Route Planner for Multi-AUV Accessing Fixed Nodes in Uncertain Ocean Environments
}

\author{Jiaxin Zhang, Meiqin Liu$^{\dag}$, Senlin Zhang, Ronghao Zheng, Shanling Dong 
	\thanks{All the authors are with the College of Electrical Engineering, Zhejiang University, Hangzhou, China. Emails: {\tt\small \{zhangjiaxin, liumeiqin, slzhang, rzheng, shanlingdong\}@zju.edu.cn}}%
	\thanks{Jiaxin Zhang, Senlin Zhang, Ronghao Zheng, and Shanling Dong are also with the National Key Laboratory of Industrial Control Technology, Zhejiang University, Hangzhou 310027, China.}
	\thanks{Meiqin Liu is also with the Institute of Artificial Intelligence and Robotics, Xi'an Jiaotong University, Xi'an, China.}
	\thanks{$^{\dag}$Corresponding author}
	\thanks{This work was supported by the National Natural Science Foundation of China under Grants U1909206 and 62173299, the Zhejiang Provincial Natural Science Foundation of China under Grant LZ23F030006, the Joint Fund of Ministry of Education for Pre-research of Equipment under Grant 8091B022147, and the Fundamental Research Funds for the Central Universities under Grant xtr072022001.}}

\begin{document}

\maketitle
\thispagestyle{empty}
\pagestyle{empty}

\begin{abstract}
This article introduces a five-tiered route planner for accessing multiple nodes with multiple autonomous underwater vehicles (AUVs) that enables efficient task completion in stochastic ocean environments. 
First, the pre-planning tier solves the single-AUV routing problem to find the optimal giant route (GR), estimates the number of required AUVs based on GR segmentation, and allocates nodes for each AUV to access. Second, the route planning tier plans individual routes for each AUV. During navigation, the path planning tier provides each AUV with physical paths between any two points, while the actuation tier is responsible for path tracking and obstacle avoidance. Finally, in the stochastic ocean environment, deviations from the initial plan may occur, thus, an auction-based coordination tier drives online task coordination among AUVs in a distributed manner.
Simulation experiments are conducted in multiple different scenarios to test the performance of the proposed planner, and the promising results show that the proposed method reduces AUV usage by $7.5\%$ compared with the existing methods. When using the same number of AUVs, the fleet equipped with the proposed planner achieves a $6.2\%$ improvement in average task completion rate.
\end{abstract}

\section{INTRODUCTION}

Human activities and industrial production have been proven to have severe impacts on the health of the ocean. Marine environmental disasters such as red tides, oil spills, and storm surges further hinder the sustainable development of the ocean \cite{willis2022cleaner}. Therefore, effective monitoring of the marine environment is crucial for protecting marine ecology and promoting  long-term development of the marine economy.

Monitoring the marine environment entails tracking physical, chemical, and biological parameters such as temperature, dissolved oxygen, and chlorophyll density \cite{thushari2020plastic}. Traditionally, researchers sail out to collect water samples from specific locations, which is a highly costly and time-consuming approach with poor timeliness \cite{Clark2019}. With advances in sensor technology, deploying buoys and other devices carrying sensors for in-situ monitoring has become the preferred method \cite{mcpartlin2017biosensors}. However, the lack of three-dimensional ocean monitoring and data recovery problems still limit the implementation of marine monitoring activities. To address this, researchers have proposed underwater sensor networks (USN) composed of environmental parameter sensors for seawater sampling \cite{demetillo2019system}, significantly improving the efficiency of marine monitoring. 

Long-term monitoring of the marine environment often relies on fixed sensor nodes deployed in the interesting ocean region. However, because these nodes are located underwater, it is often infeasible to transmit data through electromagnetic waves, and there are usually few or no cables available for information transmission \cite{Gussen2016}. Thus, using autonomous underwater vehicles (AUVs) to gather environmental monitoring data from these nodes has emerged as the most popular approach \cite{Bresciani2021}. Typically, multiple monitoring nodes are deployed in certain region, creating the challenge of how an AUV can plan its multi-node access route in a highly uncertain ocean environment to maximize the collection of monitoring data during each voyage \cite{Wei2022}. 

In \cite{dunbabin2009autonomous}, the problem of retrieving data from underwater sensor nodes using AUV was modeled as a Traveling Salesman Problem (TSP) and a multi-layered path planner was designed, but the study may be limited by not fully considering the energy limitation of an AUV. Similarly, \cite{Wei2022} proposed a multi-layered path planning framework to guide AUVs' node access activities. The navigation unit planned the AUV's physical path while the decision unit determined the AUV's access order of the nodes. However, it was also assumed that the AUV's energy was sufficient to traverse all sensor nodes in the network. Unlike the TSP problem which targets complete node access, AUVs often struggle to cover all sensor nodes in the task area due to battery energy constraints. Therefore, the problem of AUV accessing multiple nodes can be viewed as a combination of TSP and Knapsack Problems (KP) \cite{MahmoudZadeh2018A}. To maximize the total profit of a single voyage given a finite energy budget, the AUV needs to consider both how to choose nodes and how to reduce the total time spent on accessing the same group of nodes. This problem is also known as the Orienteering Problem (OP) \cite{CHOU2021A}, which has been extensively discussed in fields such as unmanned aerial vehicle mission planning \cite{Dorling2017} and tourism planning \cite{chen2020electric}.

Compared with general application scenarios, such as the drone OP problem \cite{Lan2021}, the marine environment introduces unique characteristics for AUV's multi-node access route planning. Prior to executing the mission, the AUV relies on a certain two-point path planning algorithm at the lower level to calculate the expected cost for each sub-path. Unfortunately, due to the existence of uncertain ocean current and obstacles, the actual cost associated with navigating each sub-path is unpredictable and may differ significantly from its expected value\cite{Li2023Three}. To address this issue, \cite{Zhang2022Robust} proposed a route evaluation method based on the probability density sampling of local path costs, which effectively improved the robustness of planning results in stochastic environments.

In many situations, it may be challenging for a single AUV to adequately cover a designated region that contains multiple fixed nodes. Consequently, utilizing a multi-AUV system to carry out the node data collection mission is a logical solution \cite{Yu2023traversal}. One practical way is launching an AUV repeatedly to simulate a multi-AUV system \cite{shi2022cooperative}, but it may limit the flexibility due to a lack of online cooperation. Additionally, the cost associated with recovering, recharging, and redeploying AUVs can significantly increase time and human resource consumption. Therefore, the most promising research direction for a multi-AUV system focuses on multi-AUV task allocation and coordination\cite{Wang2022task}.

Task allocation is a critical aspect of multi-AUV systems. In order to achieve this goal, \cite{Sun2022An} employed Voronoi diagrams to allocate tasks among AUVs. Besides that, \cite{Abbasi2022Cooperative} and \cite{mahmoudzadeh2019uuv} allocated tasks for individuals within multi-robot marine systems using $k$-means clustering and agglomerative hierarchical clustering (AHC) method, respectively.
Although these approaches are feasible and effective, when nodes are unevenly distributed, a significant amount of subsequent adjustment is often necessary to ensure fairness in task allocation. For example, \cite{Abbasi2022Cooperative} proposed a heuristic cooperative fleet method, while \cite{Han2023Early} proposed an improved contract algorithm to address this issue.
Once task allocation is completed for multiple AUVs and their designated nodes, the typical next step is to simplify the problem into a single-AUV route planning problem for each AUV. However, existing research mainly focuses on planning routes for multi-AUV systems under static conditions, with little online collaboration among AUVs. This limits the practicality of existing algorithms in dynamic ocean environments. To tackle this challenge, \cite{Sun2022real} designed an upper-level task planning mechanism that plans routes for each individual in real-time during navigation based on global information. \cite{MahmoudZadeh2023Cooperative} developed a task coordination mechanism specifically for situations where AUVs may become stranded. Nevertheless, existing research has not fully considered the differences between planned results and actual navigation caused by the uncertain sub-path costs due to the stochastic ocean environment, which is one of the focus of this study.

\begin{figure}[htb]
	\centering  
	\includegraphics[width=\linewidth]{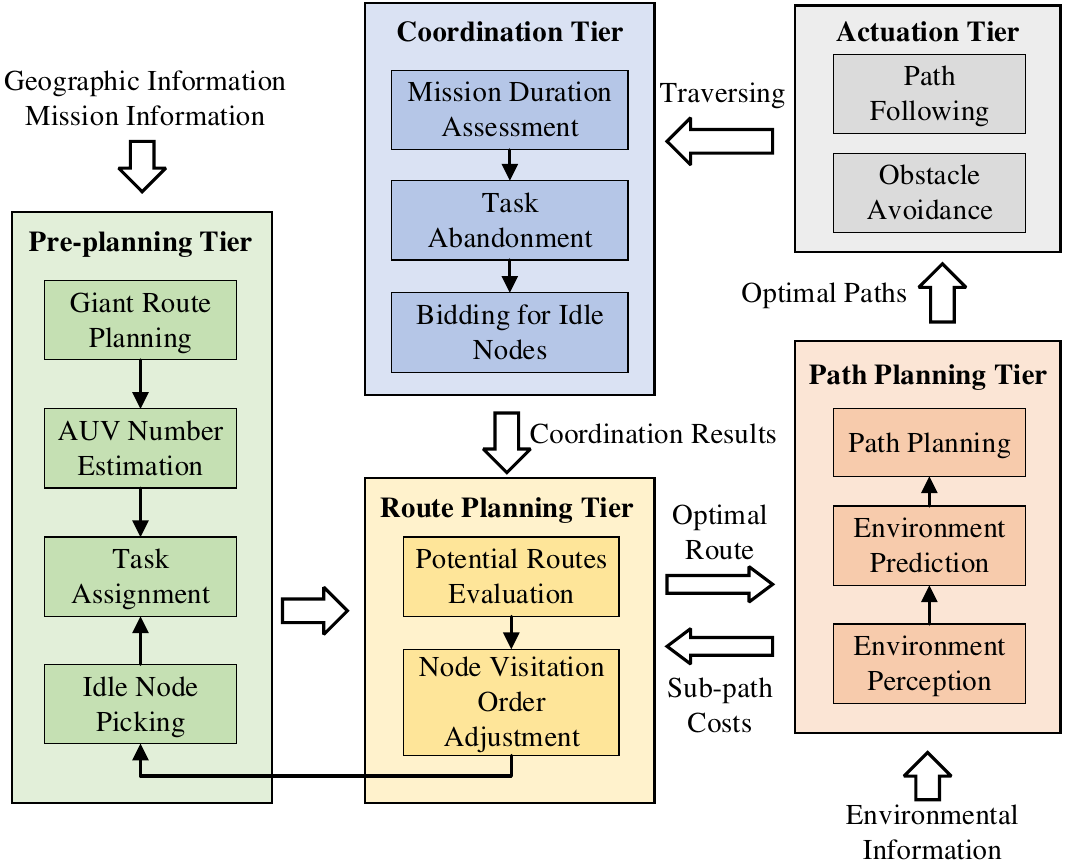}  
	\caption{Detailed structure of the five-tiered route planner.}
	\label{FiveTieredRoutePlanner}
\end{figure}

Based on the above discussion, this research aims to design a five-tiered route planner for multi-AUV accessing fixed nodes. The structure of the proposed planner is illustrated in Fig. \ref{FiveTieredRoutePlanner}. The main contributions of this study can be summarized as follows.

\begin{enumerate}[a)]
	\item A five-tiered route planner is proposed with the aim of maximizing the amount of node data collected by a multi-AUV system in a single voyage. Different tiers are responsible for multi-AUV task allocation, single AUV route planning, point-to-point path planning, path execution, and online task coordination, respectively.
	\item A giant route (GR) segmentation-based pre-planning tier is proposed, which addresses the number of AUV estimation ignored in \cite{Sun2022An}, \cite{Abbasi2022Cooperative}, and \cite{mahmoudzadeh2019uuv}, and performs better in task allocation.
	\item Given the list of nodes to be accessed, the route planning tier generates a time-optimal route for an AUV. Meanwhile, the path planning tier calculates the optimal path between every two nodes based on environmental information. AUVs navigate along the planned path based on the execution tier.
	\item An online distributed coordination mechanism based on auction algorithm is proposed, which allows idle nodes to be included in the subsequent trajectories of AUVs during the journey. The coordination tier enables significantly higher task completion rates compared with \cite{MahmoudZadeh2023Cooperative} and \cite{Sun2022real}.
\end{enumerate}

The rest of this article is organized as follows. Section \ref{formulation} formulates the problem of multi-AUV route planning. In Section \ref{prePlanningTier}, the details of the five-tiered route planner are presented. In Section \ref{simulation}, the performance of the proposed planner is thoroughly tested and compared with existing methods. Finally, Section \ref{conclusion} concludes this article.

\section{Problem Formulation}\label{formulation}
\begin{figure}
	\centering  
	\includegraphics[width=0.9\linewidth]{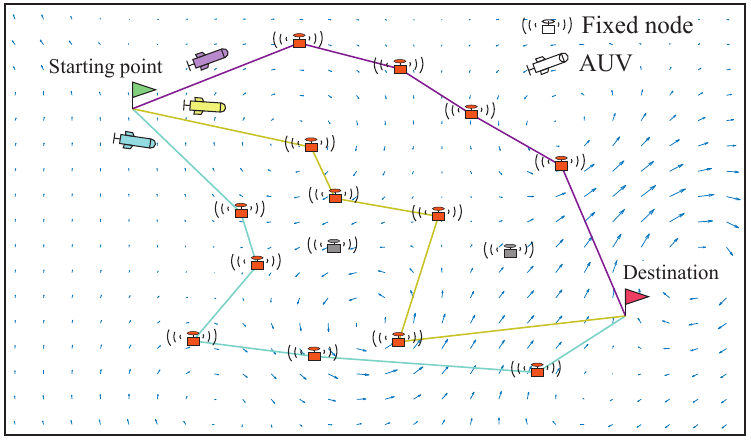}  
	\caption{Multi-AUV system for accessing fixed nodes.}
	\label{multiAUVNodeVisiting}
\end{figure}

A fleet of AUVs equipped with hydroacoustic data transmission devices is deployed to sequentially collect data from fixed sensor nodes in a specific ocean region, as shown in Fig.~\ref{multiAUVNodeVisiting}. The operational time of each AUV is limited, and the objective of the multi-AUV system is to optimize data collection within a single deployment.

\subsection{Multi-AUV Route-planning Problem}
Assuming that there are $|\bm{\mc{N}}|-2$ fixed sensor nodes in a specific ocean region, and each node $\mc{N}_i \in \bm{\mc{N}}\setminus\{\mc{N}_1,\mc{N}_{|\bm{\mc{N}}|}\}$ is associated with a positive data amount $\rho_i$, where $\bm{\mc{N}}=\{\mc{N}_1,\mc{N}_2,\dots,\mc{N}_{|\bm{\mc{N}}|}\}$ consists sensor nodes, the starting position, and the destination. The mission involves a fleet of AUVs, denoted as $\bm{\mc{A}}=\{\mc{A}_1,\mc{A}_2,\dots,\mc{A}_M\}$ with $M\geq2$, which departs from $\mc{N}_1$ and arrives at node $\mc{N}_{|\bm{\mc{N}}|}$ for recovery purposes.
The arc connecting nodes $\mc{N}_i$ and $\mc{N}_j$ is denoted by $(i,j)$. The deterministic cost for an AUV traveling from node $\mc{N}_i$ to node $\mc{N}_j$ is given by 
\begin{equation}
	\label{tij}
	t_{ij}=\frac{D(\mc{N}_i,\mc{N}_j)}{|\bm{v}_{\mc{A}}|},
\end{equation}
where $D(\mc{N}_i,\mc{N}_j)$ is the Euclidean distance between $\mc{N}_i$ and $\mc{N}_j$, and $\bm{v}_{\mc{A}}$ is the velocity of AUV. All arcs $(i,j)$ in the set of arcs $\mc{E}$ are assumed to be accessible. Then we can formulate the global route-planning problem for the multi-AUV system on the complete graph $G = \{\mc{N}, \mc{E}\}$. 

The objective of route planning is to find routes $\bm{\mc{R}}=\{\mc{R}_1,\mc{R}_2,\dots,\mc{R}_M\}$ for $\bm{\mc{A}}$, which maximize the total collected data amount represented by
\begin{equation}
	J=\sum_{m=1}^{M}\sum_{i=2}^{|\bm{\mc{N}}|-1}\rho_iy_{im},
\end{equation}
where $\rho_i$ represents the value of node $\mc{N}_i$, which is the amount of data to be collect, and $y_{im}=1$ means $\mc{N}_i$ is accessed by $\mc{A}_m$ and $y_{im}=0$ otherwise. A feasible $\bm{\mc{R}}$ should satisfy the following constraints.

Each route starts from $\mc{N}_1$ and ends at $\mc{N}_{|\bm{\mc{N}}|}$:
\begin{equation}
	\label{constraintStartEnd}
	\sum_{m=1}^{M} \sum_{j=2}^{|\bm{\mc{N}}|} x_{1 j m}=\sum_{m=1}^{M} \sum_{i=1}^{|\bm{\mc{N}|}-1} x_{i|\bm{\mc{N}}| m}=M,
\end{equation}
where $x_{ijm}=1$ indicates that $(i,j)$ is concluded in $\mc{R}_m$ and $x_{ijm}=0$ otherwise.
Each $\mc{N}_k\in\bm{\mc{N}}$ should be only accessed at most once, i. e.,
\begin{equation}
	\label{constraintVisitOnce}
	\sum_{m=1}^{M} y_{k m} \leq 1 
\end{equation}
holds for $\forall k=2, \dots,|\bm{\mc{N}}|-1$.
Meanwhile, the connectivity of each route should be ensured:
\begin{equation}
	\label{constraintConnectivity}
	\sum_{i=1}^{|\bm{\mc{N}}|-1} x_{i k m}=\sum_{j=2}^{|\bm{\mc{N}}|} x_{k j m}=y_{k m}.
\end{equation}
An AUV is powered by its battery, so the sailing time for all $\mc{A}_m\in\bm{\mc{A}}$ is constrained by
\begin{equation}
	\label{constraintTmax}
	\sum_{i=1}^{|\bm{\mc{N}}|-1} \sum_{j=2}^{|\bm{\mc{N}}|} (t_{ij}+t_j) x_{i j m} \leq T_{\max,m},
\end{equation}
where $t_j\propto \rho_j$ is the data collecting time of $\mc{N}_j$ and $T_{\max,m}$ is the maximal traveling time of $\mc{A}_m$. 
Besides, there should not exist any subtours in each route \cite{vansteenwegen2019orienteering}, that is, for all $\mc{N}_i,\mc{N}_j\in\bm{\mc{N}}$ and $\mc{A}_m\in\bm{\mc{A}}$,
\begin{equation}
	\label{constraintSubtour1}
	2 \leq u_{i m} \leq|\bm{\mc{N}}|
\end{equation}
and
\begin{equation}
	\label{constraintSubtour2}
	u_{i m}-u_{j m}+1 \leq\left(|\bm{\mc{N}}|-1\right)\left(1-x_{i j m}\right)
\end{equation}
hold, where $u_{im}$ represents that $\mc{N}_i$ is the $i$-th node to be accessed in $\mc{R}_m$.

\subsection{Environment and AUV Motion Model}
\label{environmentModel}
Ocean currents can have a significant impact on AUVs. An AUV can benefit from downstream current, which allow it to save energy, but may require additional energy to move upstream. Strong current can also potentially trap AUVs in a vortex. The behavior of ocean current is governed by the 2-D Navier-Stokes equation, which includes the vorticity term $\omega$ and the current velocity $\vec{V}=(V_x,V_y)$. The equation is given as:
\begin{equation}
	\frac{\partial\omega}{\partial t}+(\vec{V}\nabla)\omega=\nu\Delta\omega,
\end{equation}
where $\nabla$ and $\Delta$ are respectively the gradient and Laplacian operators, and $\nu$ is the viscosity of water. To simplify the mathematical modeling of ocean current, a commonly used approach is to represent the current field as a combination of Lamb vortexes \cite{garau2006auv}. Each vortex is defined by its location, core size, and circulation strength, and the vorticity of the overall field is calculated by summing the individual contributions from each vortex. The current velocity at position $\bm{r}(x,y)$ excited by a single vortex can be expressed using the following formulas:
\begin{equation}
	\begin{aligned}
		\label{vortex}
		\bm{v}_c&=(v_{c,x},v_{c,y})=f(\bm{r}_0(x_0.y_0),\Gamma,\delta),\\
		v_{c,x}&=-\Gamma\frac{y-y_0}{2\pi(\bm{r}-\bm{r}_0)^2}\left[1-e^{\frac{-(\bm{r}-\bm{r}_0)^2}{\delta^2}}\right],\\
		v_{c,y}&=\Gamma\frac{x-x_0}{2\pi(\bm{r}-\bm{r}_0)^2}\left[1-e^{\frac{-(\bm{r}-\bm{r}_0)^2}{\delta^2}}\right],
	\end{aligned}
\end{equation}
where $\bm{r_0}(x_0,y_0)$ is the vortex center, $v_{c,x}$ and $v_{c,y}$ are respectively the two horizontal components of the current velocity $\bm{v}_c$, $\Gamma$ is the vortex strength, and $\delta$ is the vortex radius.

The ground-referenced velocity of an AUV is the result of combining its propulsion speed $\bm{v}_{\mc{A}}^p$ with $\bm{v}_c$, as follows:
\begin{equation}
	\label{auvVelocity}
	\bm{v}_{\mc{A}}=\bm{v}_c+\bm{v}_{\mc{A}}^p,
\end{equation}
and the direction of $\bm{v}_{\mc{A}}$ is the same as the tangent of AUV's trajectory.
Assuming that $|\bm{v}_{\mc{A}}^p|$ is constant, and the propulsive force of the vehicle is constant as well due to the cubic relationship between speed and propulsive force, the energy consumption of the vessel is proportional to its travel time \cite{Zeng2020Exploiting}.

Eq.~\eqref{tij} provides the deterministic traveling time for an AUV move from $\mc{N}_i$ to $\mc{N}_j$. However, due to the unforeseeable ocean environment, the real traveling time is of uncertainty. Additionally, AUVs navigating in the sea will perform temporary maneuvers to avoid obstacles that appear ahead, which increases the uncertainty of actual traveling time. The real travel time from $\mc{N}_i$ to $\mc{N}_j$ is a random variable, denoted as $f(t_{ij})$, and defined by
\begin{equation}
	f(t_{ij})=t_{ij}+\Delta t_{ij}^c +\Delta t_{ij}^m,
\end{equation}
where $\Delta t_{ij}^c$ is the additional time caused by current, which follows a Gaussian distribution, and $\Delta t_{ij}^m$ is the  additional time caused by unplanned maneuvers, which occurs when unexpected obstacles appear ahead and follows a Poisson distribution \cite{Zhang2022Robust}.

\section{Multi-AUV node access route planner}
\label{prePlanningTier}
A novel five-tiered route planner is proposed for multi-AUV-based multi-node data collection needs. The five tiers of the planner play different roles in planning.


\subsection{Giant Route-based AUV Number Estimation and Task Allocation}
The multi-AUV system is introduced to address the insufficient capabilities of a single AUV. When facing multiple nodes to be accessed, the first issue to be resolved is how to determine the AUV number $M$. Additionally, tasks need to be assigned to each AUV in the system.
To this end, this section proposes a GR-based AUV number estimation and nodes allocation method.

Assuming there is only one AUV and disregarding the energy constraint, the AUV's optimal route
\begin{equation}
	\mc{R}_0=\mathop{\arg\min}_{\mc{R}_0}\sum_{i=1}^{|\bm{\mc{N}}|-1} \sum_{j=2}^{|\bm{\mc{N}}|} (t_{ij}+t_j) x_{i j}
\end{equation}
is referred to as the GR. In addition to the previously stated constraints Eq. \eqref{constraintStartEnd} -- \eqref{constraintSubtour2}, $\mc{R}_0$ should also satisfy an additional constraint: 
\begin{equation}
	\sum_{k=2}^{|\bm{\mc{N}}|-1} y_{k} = 1,
\end{equation}
which ensures that all the achievable nodes are connected by $\mc{R}_0$.
It is apparent that finding $\mc{R}_0$ with the minimal total length is a TSP. In this research, a genetic algorithm is used to solve it \cite{gupta2020comparison}.

\begin{figure*}
	\centering  
	\includegraphics[width=0.7\linewidth]{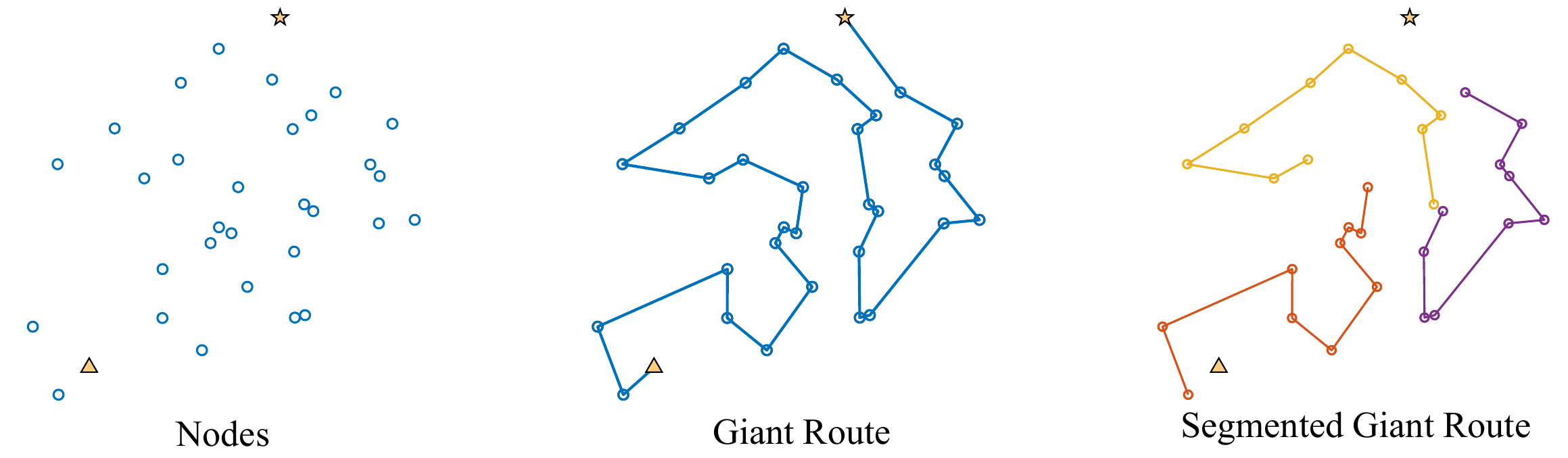}  
	\caption{Giant route segmentation-based task allocation.}
	\label{demoGiantRoute}
\end{figure*}

Due to the constraint of navigation time, it is impossible for a single AUV to complete $\mc{R}_0$. Therefore, multiple AUVs are used to jointly do this.
Obviously, a larger number of AUVs can provide more reliable completion of the task, but it also generates greater resource consumption. Therefore, it is necessary to estimate the minimal number of AUVs required to complete $\mc{R}_0$.
Based on this, let $M$ be the number of AUVs, and divide $\mc{R}_0$ into $M$ segments of sub-routes
\begin{equation}
	\mc{R}_0=\left(\mc{R}_0^{[1]},\mc{R}_0^{[2]},\dots,\mc{R}_0^{[M]}\right)
\end{equation}
with each AUV responsible for completing one sub-route. Fig. \ref{demoGiantRoute} depicts an example of this process, where $M=3$ and the nodes, start position, destination are represented by triangle, star, and circles, respectively.

\begin{assumption}
	\label{sameTmax}
	The total available travel time of each AUV is the same, i.,e.,$T_{max,m}=T_{max},\forall \mc{A}_m\in\bm{\mc{A}}$.
\end{assumption}
The time consumption for $\mc{A}_m$ is the sum of the time it takes to complete $\mc{R}_0^m$, the time it takes to move from the starting point to $\mc{R}_0^{[m]}$, and time from $\mc{R}_0^{[m]}$ to the destination. 
Based on this, and in conjunction with Assumption \ref{sameTmax}, we have
\begin{equation}
	T_0+(M-1)\tilde{t}=MT_{max},
\end{equation}
where $T_0$ is the time consumption of $\mc{R}_0$, and $\tilde{t}$ is the estimated additional time spent by an AUV in the initial and final stages of its route. Subsequently, the estimation of the number of AUVs can be obtained by
\begin{equation}
	\label{estimationM}
	M=\frac{T-\tilde{t}}{T_{max}-\tilde{t}}.
\end{equation}

After obtaining the estimated number of AUVs, it is necessary to decide where to separate $\mc{R}_0$ to obtain sub-routes. At this stage, it is not necessary to impose the total travel time limit on the AUVs because the sub-routes obtained after the division can still be further optimized. Based on the principle of load balancing, the problem to be solved is to minimize the time consumption difference:
\begin{equation}
	\label{segmentation}
	\bm{\mc{Q}}=\mathop{\arg\min}_{\bm{\mc{Q}}\in\mc{R}_0}\sum_{i=1}^{M}\left(T(\mc{R}_i)-\frac{1}{M}\sum_{j=1}^{M}T(\mc{R}_j)\right),
\end{equation}
where $\bm{\mc{Q}}=\left\{\mc{Q}_1,\mc{Q}_2,\dots,\mc{Q}_{M-1}\right\}$ is the set of segmentation points, $T(\mc{R}_i)$ is the traveling time of $\mc{R}_i$, and $\mc{Q}_i=q$ means $\mc{R}_0$ will be segmented after the $q$-th node.

Up to this point, the $M$ sub-routes are obtained. Each AUV is assigned by a set of nodes that are connected by the corresponding sub-route. 

\subsection{Route and Path Planning for Single AUV}
Despite that $\mc{R}_0$ is optimal, it does not guarantee that a sub-route obtained after segmentation will also be the optimal route connecting certain nodes. Additionally, there is no strict adherence to the time constraint for navigation. Therefore, it is necessary to further optimize the sub-routes $\left\{\mc{R}_0^{[1]},\mc{R}_0^{[2]},\dots,\mc{R}_0^{[M]}\right\}$.

The local route optimization is a single AUV multi-node access planning problem in a stochastic environment, where the optimization target for $\mc{A}_m$ is to maximize
\begin{equation}
	\sum_{j=2}^{|\bm{\mc{N}}|-1}\rho_jy_{jm}.
\end{equation}
The problem can be carried out using the sampling-base genetic strategy genetic algorithm (S-GGARP) proposed in our previous research \cite{Zhang2022Robust}. For any $\mc{R}_0^{[m]}$, the optimization is described as
\begin{equation}
	\mc{R}^{[m]}=\mathbb{T}\left(\mc{R}_0^{[m]}\right),
\end{equation}
where $\mathbb{T}$ is the operator used for optimizing $\mc{R}_0^{[m]}$, and $\mc{R}^{[m]}$ represents the optimal route for $\mc{A}_m$. 

If sub-routes for AUVs exceed the designated time limit, some nodes may have to be skipped. The pre-planning tier will check whether any of the previously discarded nodes can be added to the routes of any AUVs, and the optimization will be repeated until the routes can no longer be optimized. 
After the above process, the fleet is prepared for deployment. Algorithm \ref{AlgorithmPreplanning} describes the proposed pre-planning method which consists of the pre-planning tier and the route planning tier using pseudocode.


\begin{algorithm}[htb]
	\caption{GR-based multi-AUV route planning before departure}
	\begin{algorithmic}[1]
		\label{AlgorithmPreplanning}
		\State \textbf{Input:} Mission region map, information of node set $\bm{\mc{N}}$, maximal travel time $T_{max}$.\\
		\State Find $\mc{R}_0$ by solving $\mc{R}_0 = \mathop{\arg\min}_{\mc{R}_0}\sum_{i=1}^{|\bm{\mc{N}}|-1} \sum_{j=2}^{|\bm{\mc{N}}|} (t_{ij}+t_j) x_{i j}$
		\State Estimate additional time $\tilde{t}=t_{1\left(\mc{N}_{\mc{R}_0^1}\right)}+t_{\left(\mc{R}_0^{|\bm{\mc{N}}|}\right)|\bm{\mc{N}}|}$
		\State Calculate AUV number $M=\frac{T-\tilde{t}}{T_{max}-\tilde{t}}$\\
		\State Find $\bm{\mc{Q}}$ by solving Eq.~\eqref{segmentation}, segment $\mc{R}_0$, and get sub-routes $\left\{\mc{R}_0^{[1]},\mc{R}_0^{[2]},\dots,\mc{R}_0^{[M]}\right\}$
		\State Initial routes $\mc{R}^{[m]}=\mc{R}^{[m]}_0$
		\While{Sub-routes can be further improved}
		\State $\mc{R}^{[m]}=\mathbb{T}\left(\mc{R}^{[m]}\right)$
		\For{$m=1:M$}
		\If{Exists idle node can be picked up by $\mc{A}_m$}
		\State Add idle node into $\mc{R}^{[m]}$
		\EndIf
		\EndFor
		\EndWhile
		\Return Optimal initial routes $\left\{\mc{R}^{[1]},\mc{R}^{[2]},\dots,\mc{R}^{[M]}\right\}$
	\end{algorithmic}
\end{algorithm}

Given a planned route, the path planning tier provides an AUV with a physical path to follow between any two nodes. A feasible path from $\mc{N}_i$ to $\mc{N}_j$ should satisfies 
\begin{equation}
	\mc{P}_{ij}\cap \{\bm{\mc{O}},\bm{\mc{L}}\}=\varnothing,
\end{equation}
where $\bm{\mc{O}}$ is obstacles in the operation region and $\bm{\mc{L}}$ is land, which means there exists no possible collision.

Discretizing $\mc{P}_{ij}$ into $n-1$ segments, the time consumption of the path, as well as the optimal target to be minimized, is calculated by
\begin{equation}
	\label{optimizationfunction}
	t_{ij}=\int t\mathrm{d}\mathcal{P}_{ij} \approx\sum_{k=1}^{n}t_k=\sum_{k=1}^{n}\frac{|\bm{p}_{k+1}-\bm{p}_{k}|}{|\bm{v}_{\mc{A}|}},
\end{equation}
where $\bm{v}_{\mc{A}}$ is the ground-referenced velocity of the AUV in the segment path $(\bm{p}_k,\bm{p}_{k+1})$, and $\bm{p}_k$ and $\bm{p}_{k+1}$ are adjacent way points in the path. The ground-referenced velocity of an AUV $\bm{v}_{\mc{A}}$ at any position of its path is defined by Eq.~\eqref{auvVelocity}. Eq.~\eqref{optimizationfunction} suggests that the local current influences the time cost of $(\bm{p}_k,\bm{p}_{k+1})$. Based on horizontal acoustic doppler current profiler (H-ADCP), an AUV is able to infer the 2D structure of the ocean current field in front of it \cite{garau2006auv}. So it is possible for an AUV to take advantage of the ocean current to minimize its travel time. 

\begin{figure}
	\centering  
	\includegraphics[width=0.8\linewidth]{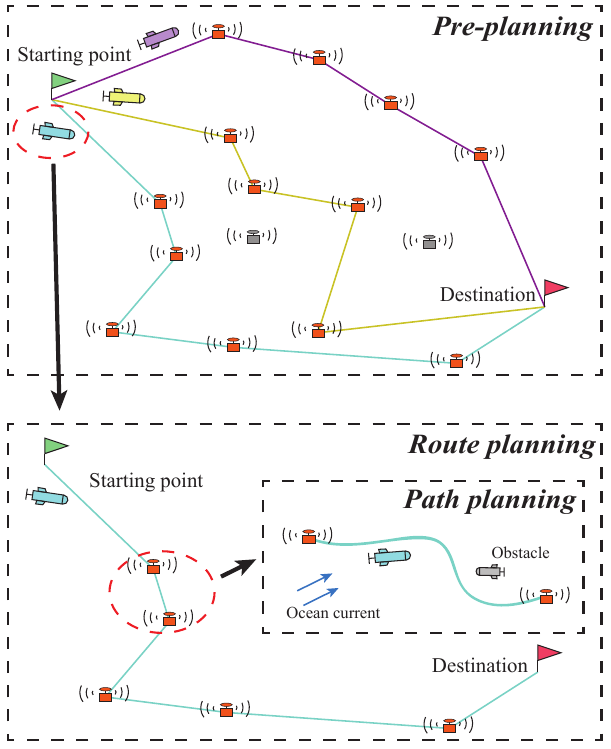}  
	\caption{Relationship among pre-planning, route planning, and path planning.}
	\label{relationshipPlanning}
\end{figure}

Consequently, the path planning tier's duty is to solve the following optimization problem:
\begin{equation}
	\begin{aligned}
		\mc{P}_{ij}^*=\mathop{\arg\min}_{\mc{P}_{ij}^*\in\bm{\mc{P}_{ij}}}t_{ij}=\mathop{\arg\min}_{\mc{P}_{ij}^*\in\bm{\mc{P}_{ij}}}\sum_{k=1}^{n}\frac{|\bm{p}_{k+1}-\bm{p}_{k}|}{|\bm{v}_{g,k}|}&\\
		s.t.\ \bm{p}_0=\mc{N}_i,\bm{p}_n=\mc{N}_j,\bm{\mathcal{M}}_{AUV}=0&
	\end{aligned}
\end{equation}
where $\bm{\mathcal{M}}_{AUV}$ is AUV's kinematic model, which means sharp curves are undesired. Research \cite{Zhang2022AUV} designed a differential evolution-based path planner for an AUV working in complex environment, and this study embedded the planner as the path planning tier into the proposed five-tiered route planner, providing AUVs with two-point path planning services. The relationship among pre-planning, route planning, and path planning is depicted in Fig.~\ref{relationshipPlanning}.

During the navigation, the actuation tier leads an AUV to track the path provided by the path planning tier based on the control system. If obstacles are detected ahead, the maneuver program is invoked to perform maneuvers avoid them. Assuming that the AUV utilizes a consistent maneuvering strategy during the movement, each maneuver will result in the same additional travel time $\Delta t^m$ \cite{Zhang2022Robust}. The actuation tier, which directly interacts with the environment, determines the actual travel time for the AUV to navigate from $\mc{N}_i$ to $\mc{N}_j$. The AUV monitors the real-time difference between its navigation status and the plan, triggering online coordination at appropriate times.

\subsection{Auction-based Online Coordination}
If the fleet operates in a static environment, the results provided by aforementioned planning tiers are sufficient. However, the uncertain ocean environment often hinders AUVs from completing the planned routes. 


While the path planning tier enhances the local motion efficiency of AUVs, it is still impossible to predict ocean currents and obstacles with complete accuracy. An AUV may spend more than $t_{ij}$ moving from $\mc{N}_i$ to $\mc{N}_j$. When the additional time spent is too much, the AUV will be unable to complete the planned journey with the remaining energy, and it must take measures to avoid being unable to reach the destination. Specifically, when $\mc{A}_m$ finds that
\begin{equation}
	\label{timeViolation}
	\hat{t}\left(\mc{A}_m,\mc{N}_{|\bm{\mc{N}}|}\right)>T_{max}-t\left(\mc{N}_1,\mc{A}_m\right),
\end{equation}
where $t\left(\mc{N}_1,\mc{A}_m\right)$ represents the time taken by $\mc{A}_m$ to travel from its starting position to its current position, and $\hat{t}\left(\mc{A}_m,\mc{N}_{|\bm{\mc{N}}|}\right)$ refers to the estimated time required to complete the remaining planned route, some nodes to be accessed will be discarded. Assume that route $\mc{R}^{[m]}=\left(\mc{N}_{\mc{R}^{[m]}}^1,\mc{N}_{\mc{R}^{[m]}}^2,\dots,\mc{N}_{\mc{R}^{[m]}}^{|\mc{R}^{[m]}|}\right)$ is the remaining route for $\mc{A}_m$, which connects $|\mc{R}^{[m]}|$ nodes. Define the effectiveness-cost ratio of $\mc{N}_{\mc{R}^{[m]}}^i$ as
\begin{equation}
	\label{costEffectiveness}
	\phi_{\mc{R}^{[m]}}^i=\frac{\rho_{\mc{N}_{\mc{R}^{[m]}}^i}}{\Delta t_{\mc{R}^{[m]}}^i},
\end{equation}
where $\Delta t_{\mc{R}^{[m]}}^i$ is the saved travel time by deleting $\mc{N}_{\mc{R}^{[m]}}^i$ in $\mc{R}^{[m]}$. Whenever $\mc{A}_m$ detects a violation in time as defined by Eq.~\eqref{timeViolation}, it will systematically discard nodes to be accessed starting from the lowest $\phi_{\mc{R}^{[m]}}^i$, until the total sailing time constraint is satisfied again.

This process leads to a lower actual task completion rate than what is anticipated for the AUV fleet. However, the impact of uncertain ocean environments on AUV navigation is not always adverse. As described in Section \ref{environmentModel}, because of the presence of ocean currents, there are instances when an AUV's actual travel time is shorter than expected. This creates opportunities for the AUV to add nodes that are currently idle into its subsequent route.

This problem is not complicated when there is only one AUV having the capacity to include available nodes in its subsequent route. Similar to Eq.~\eqref{costEffectiveness}, define the effectiveness-cost ratio of including the available node $\mc{N}_j$ in route $\mc{R}^{[m]}$ as
\begin{equation}
	\psi_{\mc{R}^{[m]}(i)}^j=\frac{\rho_{j}}{\Delta t_{\mc{R}^{[m]}(i)}^{j}},
\end{equation}
where $\Delta t_{\mc{R}^{[m]}(i)}^{j}$ is the additional travel time generated by inserting $\mc{N}_j$ at the $i$-th position of $\mc{R}^{[m]}$. The efficiency-cost ratio of including $\mc{N}_j$ in $\mc{R}^{[m]}$, in order to access more nodes with the minimum amount of additional energy expenditure, is defined as:
\begin{equation}
	\psi_{\mc{R}^{[m]}}^j=\max\left\{\psi_{\mc{R}^{[m]}(i)}^j\right\},
\end{equation} 
and if inserting $\mc{N}_j$ into $\mc{R}^{[m]}$ at any position would exceed the traveling time limitation, then $\psi_{\mc{R}^{[m]}}^j=0$. When there is sufficient endurance for $\mc{A}_m$, all $\mc{N}_j \notin \bm{\mc{R}}$ are sequentially attempted to be inserted into $\mc{R}^{[m]}$ in decreasing order of $\psi_{\mc{R}^{[m]}}^j$.

Since the system is composed of multiple AUVs that operate independently, there are chances that several AUVs may simultaneously detect and try to collect idle nodes. The problem at hand is to determine the most effective allocation of multiple nodes to multiple AUVs, with the goal of maximizing the total amount of data collection.
Assuming that there are $J$ idle nodes to be picked up, now consider the possibility of matching the $J$ nodes with the $M$ AUV through a market mechanism, viewing each AUV as an economic agent acting in its own best interest. 

When $\mc{A}_m$ intends to include node $\mc{N}_j$ in its future navigation plan, it incurs a virtual cost represented by $b_{mj}$. The AUV wants to be assigned to the node $\mc{N}_{j_m}$ with maximal profit, that is,
\begin{equation}
	\psi_{\mc{R}^{[m]}}^{j_m}-b_{mj_m}=\max\left\{\psi_{\mc{R}^{[m]}}^{j}-b_{mj}\right\},
\end{equation}
and naturally, the node brings $\mc{A}_m$ optimal profit is
\begin{equation}
	\mc{N}_{j_m}=\mathop{\arg\max}_{j=1,2,\dots,J}\left\{\psi_{\mc{R}^{[m]}}^{j}-b_{mj}\right\}.
\end{equation}

Once nodes that are available for picking up appear, AUVs within the system begin to compete by placing bids on these nodes. The bidding process begins with all AUVs placing a starting bid of $b_{mj}=0$ on their target nodes. Then, in each round of bidding, all AUVs dynamically raise their bids by the increment $\epsilon$ until either all nodes are allocated or no further changes occur in the bids placed by the AUVs. Since the virtual bids made by AUVs do not have any real cost associated with them, it is practical to set $\epsilon$ of each AUV in each round to a higher value within permissible limits. This can help expedite the convergence of the algorithm and reduce the time taken for node allocation.

\begin{figure}[htb]
	\centering
	\begin{tabular}{c}
		\includegraphics[width=0.9\linewidth]{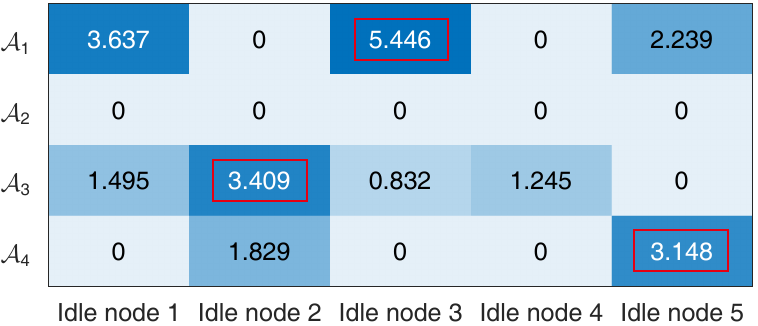}\\
		{\footnotesize\sf (a) First round of bidding} \\
		\includegraphics[width=0.9\linewidth]{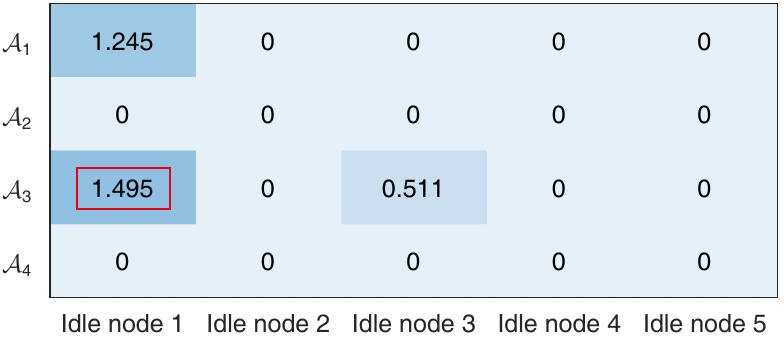}\\
		{\footnotesize\sf (b) Second round of bidding} \\
	\end{tabular}
	\caption{Auction-based idle nodes assignment}
	\label{AuctionHeatmap}
\end{figure}

Fig.~\ref{AuctionHeatmap} shows a typical auction process consisting of two rounds of bidding, where the bids of each AUV for each node are presented, and the winning bids are marked by red boxes. After the coordination, four out of five idle nodes are allocated to three AUVs, while one node remains idle and one AUV is not assigned any new nodes.
Algorithm~\ref{nodePickUp} provides a pseudocode description of the idle node allocation process that occurs during the navigation.

\begin{algorithm}[htb]
	\caption{Auction-based Idle Nodes Assignment}
	\label{nodePickUp}
	\begin{algorithmic}[1]
		\State \textbf{Input:} Idle node set $\bm{\mc{N}}_{idle}$, elapsed time $t_c$, maximal travel time $T_{max}$, AUVs' positions $\{\bm{x}_m\}_{m=1,\dots,M}$, subsequential routes $\left\{\mc{R}^{[m]}\right\}_{m=1,\dots,M}$.
		\State \textbf{Initial:} $b_{mj}=0,\forall \mc{A}_m\in\bm{\mc{A}},\forall\mc{N}_j\in\bm{\mc{N}}_{idle}$. \label{initial}
		\For{$m=1:M$}
		\For{$j=1:J$}
		\For{$i=1:|\mc{R}^{[m]}|$}
		\If{$T_{max}-t_c>\Delta t_{\mc{R}^{[m]}(i)}^{j}$}
		\State $\psi_{\mc{R}^{[m]}(i)}^j=\rho_{j}/\Delta t_{\mc{R}^{[m]}(i)}^{j}$
		\Else 
		\State $\psi_{\mc{R}^{[m]}(i)}^j=0$
		\EndIf
		\EndFor
		\State $\psi_{\mc{R}^{[m]}}^j=\max\left\{\psi_{\mc{R}^{[m]}(i)}^j\right\}$
		\State Best node $\mc{N}_{j_m}=\mathop{\arg\max}_{j=1,2,\dots,J}\left\{\psi_{\mc{R}^{[m]}}^{j}-b_{mj}\right\}$
		\EndFor
		\EndFor
		\For{$j=1:J$}
		\For{$m=1:M$}
		\State Update $b_{mj}=b_{mj}+\epsilon_{mj}$
		\EndFor
		\If{$\forall m,\epsilon_{mj}=0\ \&\&\ \exists b_{mj}>0$}
		\State $\mc{A}_{win}=\mathop{\arg\max}_{\mc{A}_m\in\bm{\mc{A}}} b_{mj}$
		\State Add $\mc{N}_j$ to $\mc{R}^{[win]}$
		\State $\bm{\mc{N}_{idle}}=\bm{\mc{N}_{idle}}\backslash\mc{N}_j$
		\State Jump to line \ref{initial}
		\EndIf
		\EndFor	
	\end{algorithmic}
\end{algorithm}

\label{planner}

\section{Simulations}
\label{simulation}
To verify the effectiveness of the proposed five-tiered route planner, numerical simulations are conducted on a computer equipped with an Intel i7-8700@3.20-GHz CPU using MATLAB 2022b. In this section, detailed description of the simulation scenarios and parameters are provided, and the performance of the proposed planner is compared with existing methods in Monte Carlo (MC) tests.

\subsection{Experimental Settings}
In this research, an AUV fleet with uncertain AUV number is deployed to download data collected by fixed nodes scattered in a $10km\times10km$ ocean region. In the system, every AUV is released from the same location and travels for a certain duration to collect data before reaching a predetermined point where it is recovered. Every AUV moves at a velocity of $1m/s$ and is able to operate continuously for up to $5$ hours on its battery. 

Within the mission area, there exist 60 fixed nodes that need to be accessed, and their locations are randomly generated for each individual experiment. 
The amount of data stored in node $\mc{N}_j\in\bm{\mc{N}}$ is normalized to $\rho_j \in (0,1]$. The time required for an AUV to collect data from $\mc{N}_j$ is as follows:
\begin{equation}
	t_j=t_0+20\rho_{j}\ s,
\end{equation}
where $t_0=5\ s$ is the fixed communication delay.

In the operation region, there exists a current field which consists of $20$ vortexes defined by Eq.~\eqref{vortex}. The parameters are set as follows: $\Gamma_i\sim N(0,50), \delta_i\sim N(80,100)\ (i=1,2,\dots,20)$, and the center of each vortex randomly lands within the region. 

In order to assess the degree of task completion, the task completion rate in one deployment is defined as the ratio of the sensor data collected to the total amount of sensor data that can be collected, expressed as follows
\begin{equation}
	\theta=\frac{\sum_{m=1}^{M}\sum_{i=2}^{|\bm{\mc{N}}|-1}\rho_iy_{im}}{\sum_{N_i\in\bm{N}}\rho_i}.
\end{equation}

\subsection{Planning in Deterministic Environments}
When the stochastic ocean environment is not taken into consideration, the operational result is expected to match that of the pre-planning phase.
Therefore, the pre-planner which offers estimation of AUV number and route pre-planning services is tested in this part. Unlike existing task allocation methods, the proposed planner does not require a designated number of AUV. 

\begin{figure*}[htb]
	\centering
	\includegraphics[width=0.9\linewidth]{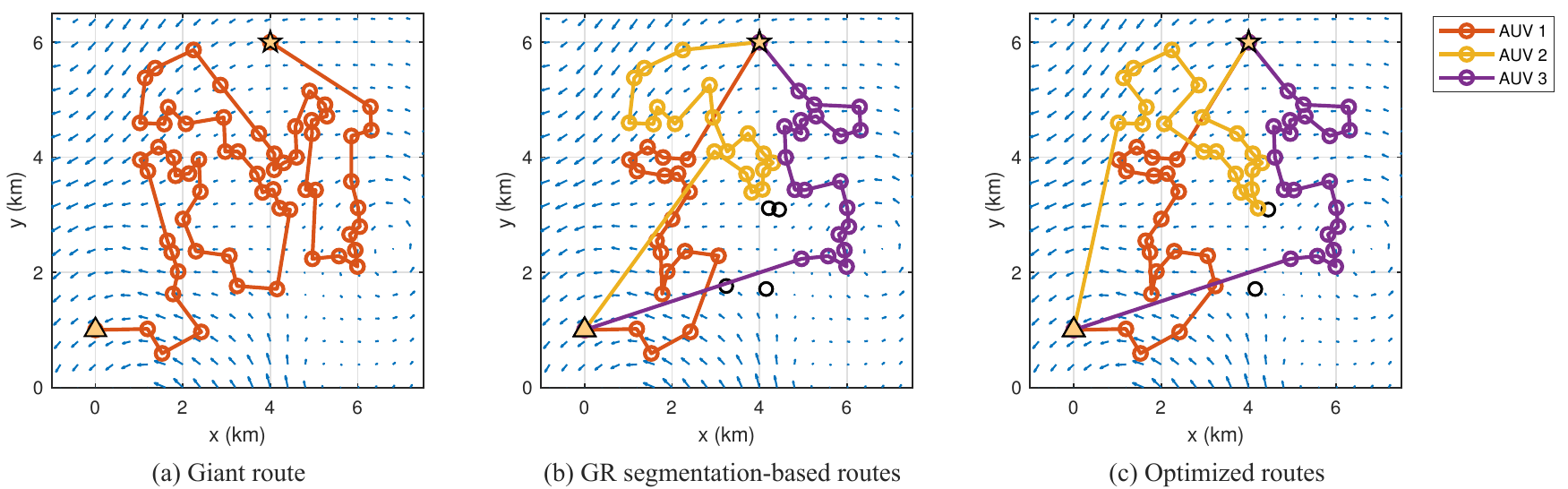}  
	\caption{Entire process of the pre-planning.}
	\label{preplanning}
\end{figure*}

This study evaluates the performance of the GR-based task allocation and route optimization process, and Fig.~\ref{preplanning} shows a typical result of the pre-planning process when $M=3$.
Initially, the GR is generated to connect all the candidate nodes, as shown in Fig.~\ref{preplanning}(a). Then, the AUV number is estimated using Eq.~\eqref{estimationM}, and the GR is segmented based on load balancing principles.
Subsequently, each sub-route is assigned to one AUV in the team, and nodes may be deleted from the routes at this stage to satisfy the energy constraint defined by Eq.~\eqref{constraintTmax}, as shown in Fig.~\ref{preplanning}(b).
Finally, optimization is performed on each AUV's route, resulting in changes to the node access sequences of some routes, as shown in Fig.~\ref{preplanning}(c). Additionally, previously idle nodes are now included in the routes.

Most of the current research on multi-AUV task allocation problem focuses on the spatial relationships of nodes, such as Voronoi-based spatial partitioning \cite{Sun2022An}, k-means clustering-based task assignment \cite{Abbasi2022Cooperative}, and AHC-based task assignment \cite{mahmoudzadeh2019uuv}.
The proposed planner is tested in 20 different scenarios and compared with existing algorithms.
The proposed method introduces a mechanism for AUV number estimation, while existing algorithms typically rely on manual and empirical settings of AUV number. Therefore, the comparison experiments are conducted under two conditions:
\begin{itemize}
	\item In a single test scenario, the three existing algorithms set the number of AUVs based on the estimation results of the proposed pre-planner:
	\begin{equation}
		M_{GR}=M_{Voronoi}=M_{k-means}=M_{AHC}.
	\end{equation}
	\item The AUV numbers in the three algorithms used for comparison are empirically set to
	\begin{equation}
		M_{Voronoi}=M_{k-means}=M_{AHC}=4.
	\end{equation}
\end{itemize}

\begin{figure}[htb]
	\centering  
	\includegraphics[width=0.8\linewidth]{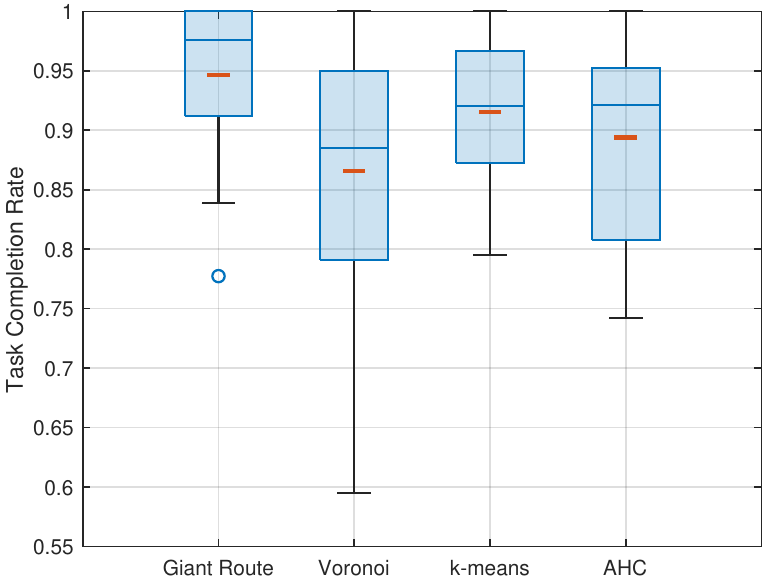}  
	\caption{Expected task completion rate given by pre-planners with consistent AUV numbers.}
	\label{PrePlanBoxchart}
\end{figure}

\begin{figure}[htb]
	\centering  
	\includegraphics[width=0.8\linewidth]{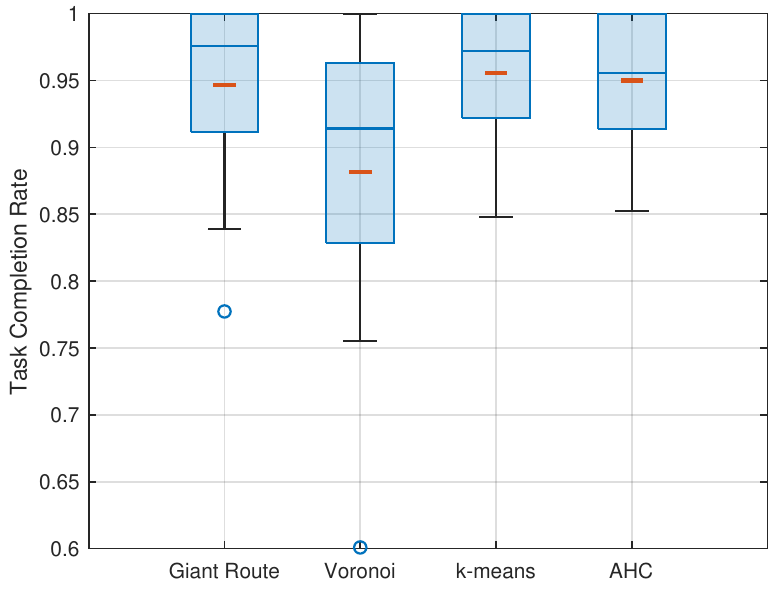}  
	\caption{Expected task completion rate given by pre-planners with inconsistent AUV numbers.}
	\label{PrePlanBoxchart2}
\end{figure}

The results of the experiments are presented by box charts in Fig.~\ref{PrePlanBoxchart} and Fig.~\ref{PrePlanBoxchart2}. The box charts presented in this study use different visual elements to represent statistical measures. Specifically, the central mark corresponds to the median value, while the central short mark indicates the average. The edges of the box indicate the 25th and 75th percentiles of the data, while the whiskers extend to the most extreme data points not considered outliers. Any outliers are plotted individually using the `o' symbol.

It is demonstrated that when the number of AUVs is consistent, the proposed planner is significantly better performance than its competitors. Specifically, its average expected task completion rate is $\theta_{GR}=0.947$, while the average task completion rates of the others are $\theta_{Voronoi}=0.866$, $\theta_{k-means}=0.916$, and $\theta_{AHC}=0.894$, respectively. Meanwhile, it is evident that the lower limit of the proposed planner's performance is also better than others.

It is common practice to manually preset the number of AUVs when using existing planning algorithms since estimating the required AUV number can be challenging. This estimation is usually based on the size of the task area and the scale of the node set. The experimental results presented in Fig.~\ref{PrePlanBoxchart2} indicate that the proposed planner achieves nearly identical results to those of the k-means and AHC-based planners, with average task completion rates of $\theta_{k-means}=0.956$ and $\theta_{AHC}=0.950$, respectively. 
However, a slightly lower expected task completion rate does not necessarily mean that the proposed planner is inferior. Because, in multiple experiments, the proposed GR-based pre-planner uses an average of $\overline{M}=3.7$ AUVs, which is $7.5\%$ lower compared with other algorithms. Therefore, the proposed planner is able to achieve comparable performance with existing methods, while incurring significantly lower operating costs.

\subsection{Planning in Stochastic Environments}
During the navigation, it is impossible for an AUV to perfectly match its expected journey due to the inability to accurately predict environmental conditions in advance. The actual travel times of AUVs often deviate from the planned ones. Thus, online coordination is necessary. The proposed five-tiered planner provides a solution for the multi-AUV route planning problem in such scenarios.

In this part, the proposed planner is tested in $20$ different scenarios. In each scenario, there exist ocean current and obstacles that can not be predicted precisely in the mission region. Meanwhile, the proposed planner is compared with planners proposed by \cite{MahmoudZadeh2023Cooperative} and \cite{Sun2022real}. 

\begin{figure*}[htb]
	\centering  
	\includegraphics[width=0.9\linewidth]{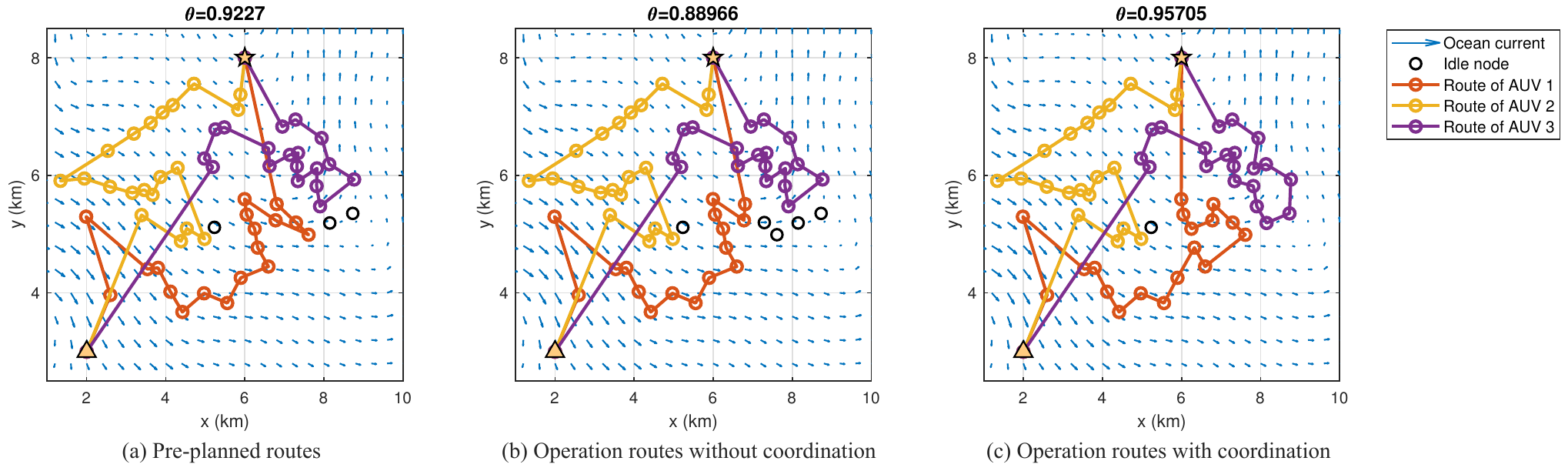}  
	\caption{The impact of the online coordination ($M=3$).}
	\label{coordination3AUV}
\end{figure*}

\begin{figure*}[htb]
	\centering  
	\includegraphics[width=0.9\linewidth]{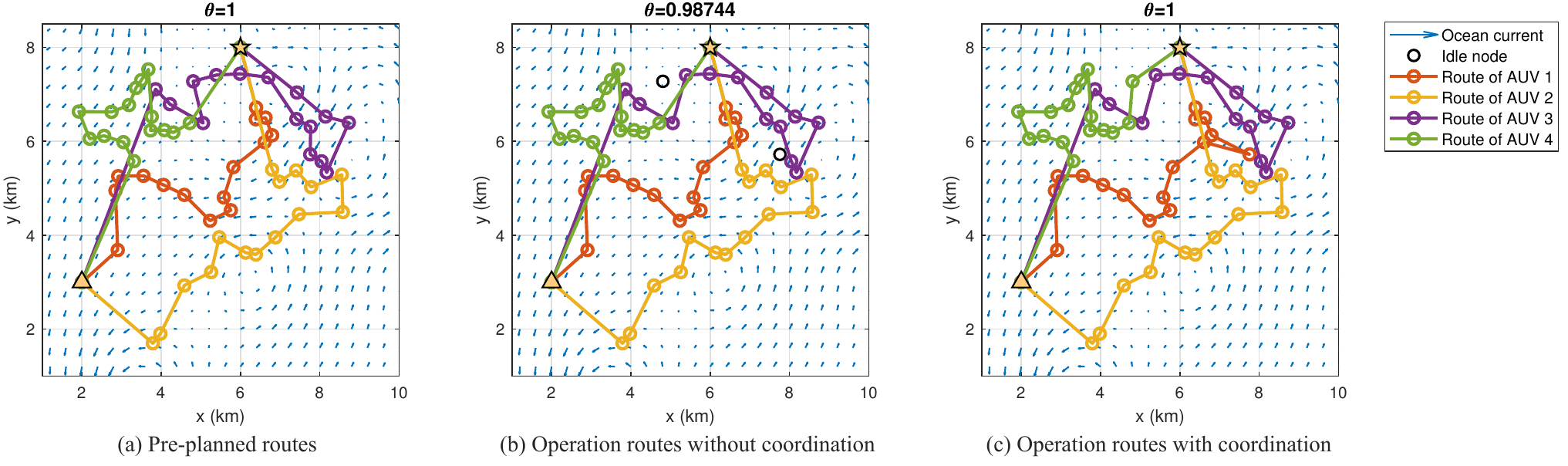}  
	\caption{The impact of the online coordination ($M=4$).}
	\label{coordination4AUV}
\end{figure*}

As previously mentioned, a larger $M$ results in a higher task completion rate, while a smaller $M$ incurs lower task costs. In some experimental scenarios, our planner selects a value of $3$ for the parameter $M$, while in other scenarios it selects a value of $4$. Fig.~\ref{coordination3AUV} and Fig.~\ref{coordination4AUV} depict typical test results for these two cases, respectively. 

In Fig.~\ref{coordination3AUV}, $3$ AUVs are deployed to gather data from fixed nodes. Fig.~\ref{coordination3AUV}(a) depicts the initial routes planned before departure by the pre-planning and route planning tiers. Fig.~\ref{coordination3AUV}(b) illustrates the operating result when the coordination tier is offline. The path planning and actuation tiers ensure that the AUVs successfully complete the mission. However, due to the complex environment, the actuation tier has to command AUVs to discard some nodes during the navigation. As a result, the actual task completion rate is significantly lower than expected. The proposed online coordination mechanism effectively suppresses the degradation of execution performance caused by random environmental factors. As shown in Fig.~\ref{coordination3AUV}(c), the multi-AUV system guided by the complete five-tiered planner achieves a task completion rate of $\theta=0.957$, which is $7.6\%$ higher than the case without coordination mechanism. 

In most cases, $3$ AUVs are not enough for the mission. An example of $M=4$ is depicted by Fig.~\ref{coordination4AUV}. Unlike the case shown in Fig.~\ref{coordination3AUV}, the AUV fleet with $4$ members is capable of covering all the nodes to be accessed in this scenario. However, the AUVs navigating through uncertain environments have to give up two nodes in order to reach their final destination, resulting in an imperfect mission completion rate. The coordinating mechanism plays a role in preventing such situations from occurring. During its mission, $\mc{A}_3$ is forced to abandon two nodes that are originally designated for it to access. Nonetheless, both of these nodes are picked up by $\mc{A}_1$ and $\mc{A}_4$, who have ample time remaining in their own mission schedules. Finally, despite the influence of the environment, the team completes their mission perfectly. 

\begin{figure*}[htb]
	\centering  
	\includegraphics[width=\linewidth]{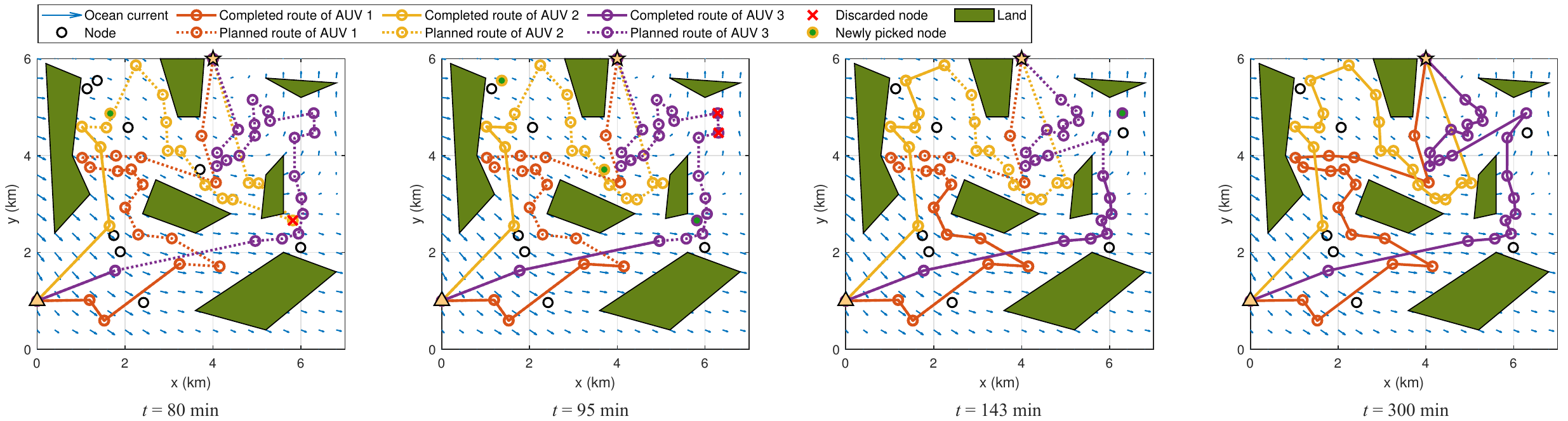}  
	\caption{Multi-AUV online task coordination during the navigation.}
	\label{demoOnlineCoordination}
\end{figure*}

Fig.~\ref{demoOnlineCoordination} offers a comprehensive visual representation of the online coordination process, with the routes planned for the AUVs prior to departure depicted as dashed lines. The fleet commences its operations along the established routes, and the parts of the routes covered are indicated with solid lines. At the time $t=80$min, $\mc{A}_2$ decides to discard a node on its subsequent route and communicates this development to its teammates, and the node is marked by a red `$\times$'. Next, the AUVs in the system initiates a bidding process for the idle nodes. Ultimately, $\mc{A}_2$ integrates a new node into its route, which is highlighted in green. The AUVs continues to monitor their task completion situation during the whole journey. The aforementioned task discarding and picking up occur multiple times. Finally, all AUVs reach the recovery station, and the mission ends.

\begin{figure}
	\centering  
	\includegraphics[width=\linewidth]{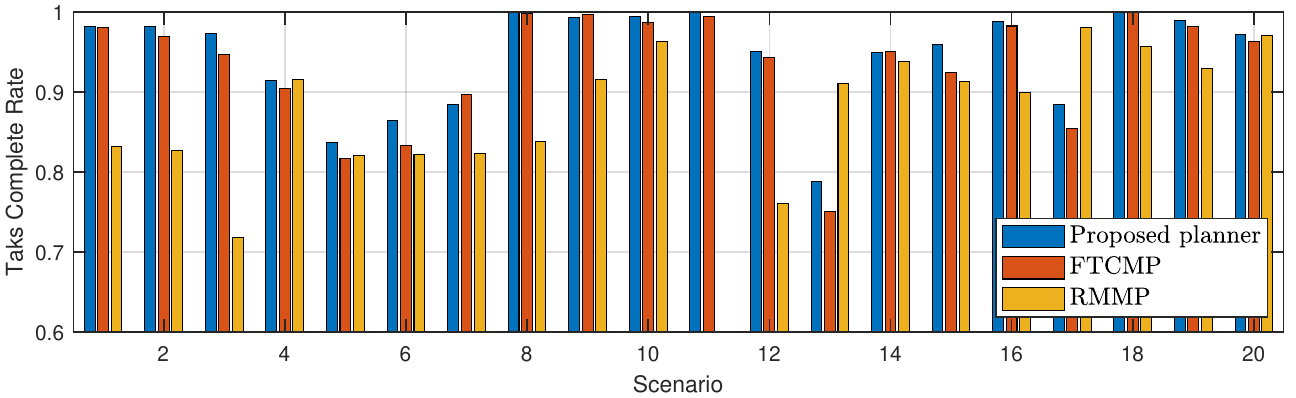}  
	\caption{Average task completion rate of different route planners in $20$ different scenarios.}
	\label{ComparisonMCTaskCompletionRate}
\end{figure}

\begin{figure}
	\centering  
	\includegraphics[width=\linewidth]{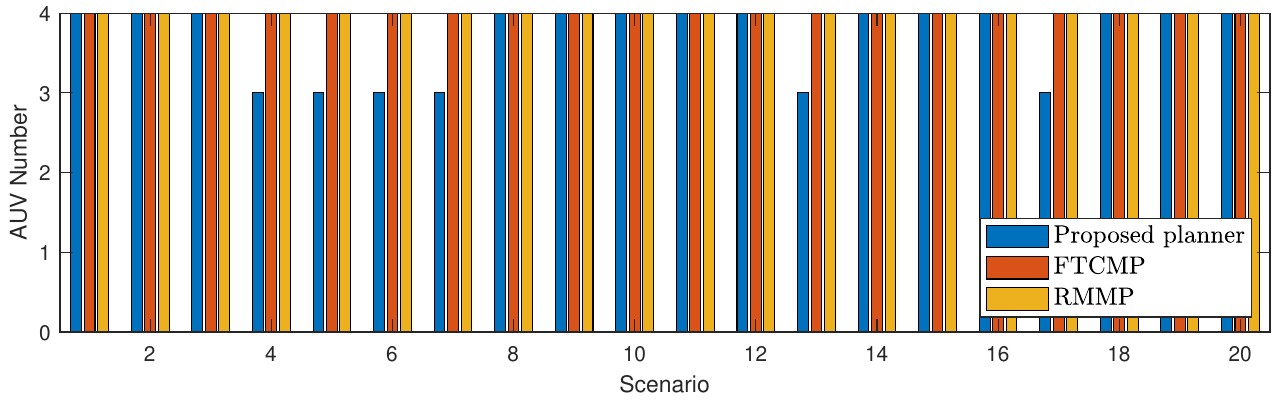}  
	\caption{AUV numbers used by different route planners in $20$ different scenarios.}
	\label{ComparisonMCAUVNumber}
\end{figure}

\begin{figure}
	\centering  
	\includegraphics[width=0.8\linewidth]{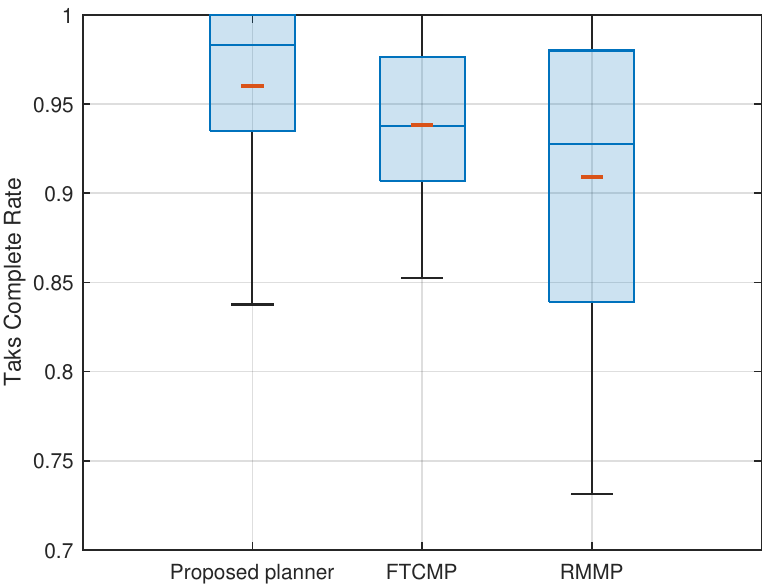}  
	\caption{Task completion rate of different route planners with empirically set numbers of AUVs (excluding the proposed planner).}
	\label{comparisonBoxchart}
\end{figure}

The proposed five-tiered route planner is tested and compared with the fault-tolerant cooperative mission planner (FTCMP) proposed by \cite{MahmoudZadeh2023Cooperative} and the real-time mission-motion planner (RMMP) proposed by \cite{Sun2022real}. In each scenario, the planners are tested $50$ times in random current fields. The average task completion rate is listed in Fig.~\ref{ComparisonMCTaskCompletionRate}. Out of the $20$ scenarios, the proposed planner demonstrates superior performance in $15$ of them, surpassing both FTCMP, which performs the best in $2$ scenarios, and RMMP, which performs the best in $3$ scenarios. Fig.~\ref{ComparisonMCAUVNumber} illustrates the suggested number of AUVs to be deployed in different scenarios by different planning algorithms. The proposed planner utilizes $4$ AUVs in $14$ out of $20$ scenarios, while in the remaining $6$ scenarios, only $3$ AUVs are deployed. In comparison, the other two planners do not provide any estimation mechanism for determining the AUV number, and thus utilize 4-AUV systems throughout the mission. The boxchart shown in Fig.~\ref{comparisonBoxchart} compares the results of the three planners. On average, the proposed planner achieves slightly better performance than FTCMP, while utilizing $7.5\%$ fewer AUVs, and significantly outperforms RMMP.

\begin{figure}
	\centering  
	\includegraphics[width=0.8\linewidth]{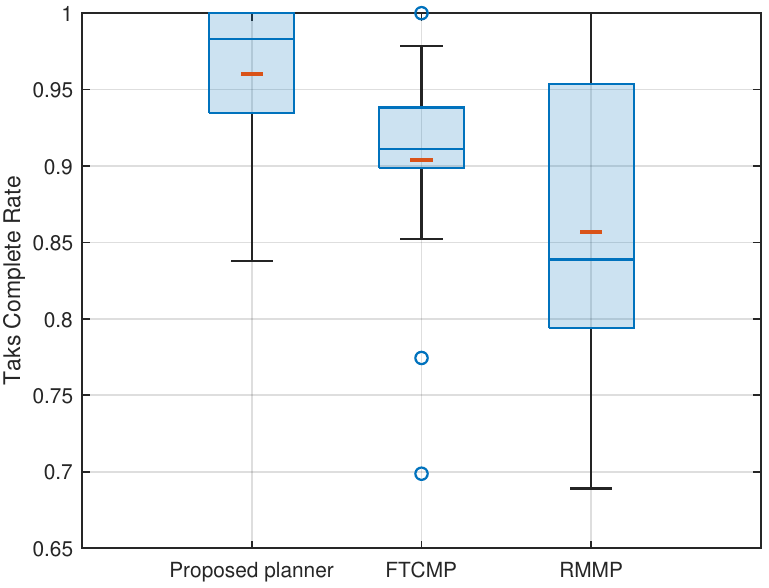}  
	\caption{Task completion rate of different route planners with the same numbers of AUVs.}
	\label{comparisonBoxchart2}
\end{figure}

Besides, the above experiments are repeated by setting the other two planners to use the same number of AUVs as the proposed planner in each test. The results are depicted in Fig.~\ref{comparisonBoxchart2}. The AUV fleet which is guided by the proposed planner achieves an average task completion rate that is $6.2\%$ higher than that of the competitors during the operation's navigation. Moreover, the lower bound of the proposed planner is significantly higher than that of the competitors.

\section{Conclusion}
\label{conclusion}
In this paper, we propose a new method to address the problem of multi-AUV-based nodes access problem in uncertain ocean environment. To this end, a novel five-tiered route planner is designed. First, the pre-planning tier estimates the required number of AUVs based on the segmentation of the optimal GR, and together with the path planning tier, plans the routes for each AUV before departure. Subsequently, under the guidance of the path planning tier, each AUV optimizes its sub-paths and performs online obstacle avoidance and path tracking based on the actuation tier. Finally, to address potential discrepancies between actual navigation and planned routes, the coordination tier leverages the auction-based idle nodes assignment algorithm to achieve online distributed task coordination. Sufficient simulation results have shown that the proposed planner effectively reduces AUV operating costs compared with existing multi-AUV task allocation methods, and realizes more efficient task allocation under equivalent conditions. In addition, compared with existing planners, the proposed five-tiered planner achieves a lower usage of AUVs or a higher task completion rate.



\bibliography{sample}
\bibliographystyle{IEEEtran}

\end{document}